\documentclass{article}
\usepackage{ijcai18}

\usepackage{times}
\usepackage{xcolor}
\usepackage{soul}
\usepackage[utf8]{inputenc}
\usepackage[small]{caption}
\usepackage[inline]{enumitem}
\usepackage{hyperref}
\usepackage{colortbl} 
\usepackage{graphicx}
\usepackage{subcaption}

\usepackage{tikz}
\def\checkmark{\tikz\fill[scale=0.25](0,.35) -- (.25,0) -- (1,.7) -- (.25,.15) -- cycle;} 
\newcommand{\textoverline}[1]{$\overline{\mbox{#1}}$}
\def\textsupersubscript#1#2{\rlap{\textsuperscript{{#1}}}\textsubscript{#2}}

\title{Empirical Analysis of Foundational Distinctions in Linked Open Data}
\author{Luigi Asprino$^1$$^,$$^2$, Valerio Basile$^3$, Paolo Ciancarini$^2$, Valentina Presutti$^1$\\ 
$^1$ STLab, ISTC-CNR (Rome, Italy) \\
$^2$ University of Bologna (Bologna, Italy) \\
$^3$ University of Turin (Turin, Italy)\\
luigi.asprino@unibo.it,
basile@di.unito.it,
paolo.ciancarini@unibo.it,
valentina.presutti@cnr.it
}

\begin{document}
\maketitle
\begin{abstract}
The Web and its Semantic extension (i.e. Linked Open Data) contain open global-scale knowledge and make it available to potentially intelligent machines that want to benefit from it. Nevertheless, most of Linked Open Data lack ontological distinctions and have sparse axiomatisation. For example, distinctions such as whether an entity is \emph{inherently} a class or an individual, or whether it is a physical object or not, are hardly expressed in the data, although they have been largely studied and formalised by foundational ontologies (e.g. DOLCE, SUMO). These distinctions belong to common sense too, which is relevant for many artificial intelligence tasks such as natural language understanding, scene recognition, and the like. There is a gap between foundational ontologies, that often formalise or are inspired by pre-existing philosophical theories and are developed with a top-down approach, and Linked Open Data that mostly derive from existing databases or crowd-based effort (e.g. DBpedia, Wikidata). We investigate whether machines can learn foundational distinctions over Linked Open Data entities, and if they match common sense. We want to answer questions such as ``does the DBpedia entity for \emph{dog} refer to a class or to an instance?". We report on a set of experiments based on machine learning and crowdsourcing that show promising results.        

\end{abstract}

\section{Common Sense and Linked Open Data}
\label{sec:intro}
\label{sec:introduction}
Common Sense Knowledge is knowledge about the world, shared by all people. Common sense reasoning is also at the core of many Artificial Intelligence (AI) tasks such as natural language understanding, object and action recognition, and behavior arbitration \cite{Davis2015}, but it is difficult to teach to those systems. Its importance was assessed back in 1989 by ~\cite{Hayes1989} who argued that AI needs a ``formalization of a sizable portion of commonsense knowledge about the everyday physical world'' (cit.), which, he says, must have three main characteristics: uniformity, density, and breadth. The Semantic Web effort has partly addressed this requirement with Linked Open Data (LOD): $\sim$150 billion linked facts\footnote{\url{http://stats.lod2.eu/}, accessed on April 25th 2018}, formally and uniformly represented in RDF and OWL, and openly available on the Web. Nevertheless, LOD still fails in addressing density (high ratio of facts about concepts) and breadth (large coverage of physical phenomena). In fact, it is very rich for domains such as geography, linguistics, life sciences and scholarly publications, as well as for cross-domain knowledge, but it mostly encodes this knowledge from an encyclopaedic perspective. The ultimate goal of our research is to  enrich LOD with common sense knowledge, going beyond the encyclopaedic view. 
%
%
We claim that an important gap to be filled towards this goal is: assessing foundational distinctions over LOD entities, that is to distinguish and formally assert whether a LOD entity inherently refers to e.g. a class or an individual, a physical object or not, a location, a social object, etc., from a common sense perspective. 

\subsection{Foundational Distinctions} High level categorial distinctions (e.g. class vs. instance) are a fundamental human cognitive ability: ``There is nothing more basic than categorization to our thought, perception, action, and speech."~\cite{Lakoff1987}. This is also why ``the organisation of objects into categories is a vital part of knowledge representation"~\cite{Russell2009}. Foundational distinctions have been theorised and modelled in foundational ontologies such as DOLCE~\cite{WONDERWEBD18} and SUMO~\cite{Pease2002} with a top-down approach, but populating and empirically validating them has been rarely addressed. In this study, we perform a set of experiments to assess \emph{whether machines can learn to perform foundational distinctions, and if they match common sense}. The former issue is investigated by applying machine learning techniques, the latter is assessed by crowdsourcing the annotation of the evaluation datasets.
We use DOLCE+DnS UltraLite (DUL)\footnote{DOLCE+DnS UltraLite (DUL) \url{http://www.ontologydesignpatterns.org/ont/dul/DUL.owl} is derived from DOLCE. DUL inherits most of DOLCE's distinctions by providing a more intuitive terminology and simplified axiomatisation. It has been widely adopted by the Semantic Web community.} as reference foundational ontology to select the targets of our experiments. We start by focusing on two very basic but diverse distinctions, which need to be addressed before approaching all the others: whether a LOD entity e.g. \texttt{dbr:Rome}\footnote{\texttt{dbr:} stands for \url{http://dbpedia.org/resource/}}, (i) inherently refers to a class or an instance, and whether it (ii) refers to a physical object or not. 
The first distinction (class vs. instance) is fundamental in formal ontology, as evidenced by upper-level ontologies (e.g. SUMO and DOLCE), and showed its practical importance in modelling and meta-modelling approaches in computer science, e.g. the class/classifier distinction in Meta Object Facility\footnote{\url{https://www.omg.org/spec/MOF/}}. It is also at the basis of LOD knowledge representation formalisms (RDF and OWL) for supporting taxonomic reasoning (e.g. inheritance). Automatically learning whether a LOD entity is a class or an instance -- from a common sense perspective  -- impacts on the behaviour of practical applications relying on LOD as common sense background knowledge. Examples include: question answering, knowledge extraction, and more broadly human-machine interaction. In fact, many LOD datasets that are commonly used for supporting these tasks (especially general purpose datasets e.g. DBpedia, Wikidata, BabelNet) only partially, and often incorrectly, assert whether their entities are classes or instances, and this has been proved to be a source of many inconsistencies and error patterns~\cite{Paulheim2015}.

%
%
%

The second distinction (physical object or not) is essential to represent the physical world. In fact, only physical objects can move in space or be the subject of axioms expressing their expected (naive) physical behaviour (e.g. gravity).   
We refer to the definition of \emph{Physical Object} provided by DUL: \emph{``Any Object that has a proper space region. The prototypical physical object has also an associated mass, but the nature of its mass can greatly vary based on the epistemological status of the object (scientifically measured, subjectively possible, imaginary)''}. 

\subsection{Contribution}
In the reminder of this paper we describe an automated 
way of making these distinctions emerge empirically. We present and discuss a set of experiments, conducted on a sample of LOD, involving manual inspection, ontology alignment, machine learning, and crowdsourcing. The obtained results are promising, and motivate us to extend them to a much larger scale, in line with a recent inspiring talk ``The Empirical Turn in Knowledge Representation''\footnote{\url{https://goo.gl/BDSGY1}} by van Harmelen, who suggests that LOD is a unique opportunity to ``observe how knowledge representations behave at very large scale''. 

In summary, the contribution of this research includes:
\begin{itemize}
\item a novel method that leverages supervised machine learning and crowdsourcing to automatically assess foundational distinctions over LOD entities (cf. Section \ref{sec:methods}), according to common sense;
\item four reusable datasets, based on a sample of DBpedia, separately annotated by experts and by the crowd with class/instance and physical object classification, for each entity (cf. Section \ref{sec:goldstandard}). The crowdsourced task designs are on their turn reusable;
\item a set of reproducible experiments targeting two foundational distinctions: \emph{class vs. instance} and \emph{physical object vs. not a physical object}, showing that machines can learn them by using a same set of features, and that they match common sense (cf. Section \ref{sec:results}).
\end{itemize}

\section{Related Work}
\label{sec:soa}
Only a few studies focus on typing entities based on \textbf{foundational distinctions}. To the best of our knowledge, our research is the first to test the hypothesis that machines can learn foundational distinctions that match common sense, by using web resources. The closest work to ours in approach and scale is~\cite{Gangemi2012}, which produced a dataset of DBpedia entities annotated with DUL classes, using ontology learning. We reuse this dataset and compare our results against it. The work by~\cite{Miller2006} addresses the problem of distinguishing classes from instances in WordNet~\cite{Fellbaum1998} synsets, through purely manual annotation. This approach is inappropriate to test our research questions due to its lack of scalability. 
%
%
Over the years, a number of \textbf{common sense knowledge bases} have been proposed for supporting diverse tasks spanning from automated reasoning to natural language processing. We decided to use
the English DBpedia\footnote{\url{http://dbpedia.org}}~\cite{Bizer2009a} in our experiments. Besides being very popular, it is the \emph{de facto} main hub of LOD, with its 4.58 million entities\footnote{\url{http://wiki.dbpedia.org/about/facts-figures}, accessed on April 25th 2018}. 
%
%
There are other resources that either target common sense, or encode potentially relevant knowledge for common sense reasoning. However they lack explicit assertions of foundational distinctions, and show very sparse coverage distribution. Two of them are relevant in this context, and we plan to investigate how to leverage them in future experiments that we plan to perform at a much larger scale:  
ConceptNet\footnote{\url{http://conceptnet5.media.mit.edu/}}~\cite{Liu2004} is a large-scale semantic network that integrates other existing resources, mostly derived from crowdsourced resources, expert-created resources, and games with a purpose. We found it unsuitable for reuse at this stage of our study: only a very small portion of it is linked to LOD, and its representation is bound to linguistic expressions instead of formalised concepts.  
%
OpenCyC\footnote{\url{http://www.opencyc.org/}
}~\cite{Lenat1995} includes an ontology and a knowledge base of common sense knowledge organised in modular theories, released as part of the long-standing CyC project. OpenCyC has been mainly developed with a top-down approach by ontology engineering experts. It is a potentially valuable resource to compare with, but it uses a proprietary representation language, which makes it hard to work with, currently. There is a plan to make it available to the scientific community as linked data, although apparently inaccessible at the time of our experiment.

\section{Automatic Classification of Foundational Distinctions}
\label{sec:methods}
We want to answer the following research questions: \emph{(RQ1) Do foundational distinctions match common sense? (RQ2) Does the (Semantic) Web provide an empirical basis for supporting foundational distinctions over LOD entities, according to common sense? and (RQ3) what ensemble of features, resources, and methods works best to make machines learn foundational distinctions over LOD entities?} 

Our objective is to test all the distinctions formalised in DUL. Nevertheless, for each of them we need to 
create a set of reference datasets (cf. Section \ref{sec:goldstandard}) in order to train and evaluate the proposed method, which requires a significant amount of work. For this reason, as anticipated in Section \ref{sec:intro}, we start focusing on two distinctions: between class and instance, and between what is a physical object and what is not. These are two of the most basic, but very diverse distinctions in knowledge representation and formal ontology. The former applies at a very high level, and is usually modelled by means of logical language primitives (e.g. \texttt{rdf:type, rdfs:subClassOf}). The latter concerns the identification of those entities that constitute the physical world, hence highly relevant and primitive as far as common sense about physics is concerned. We argue that by investigating these two distinctions, given their diverse character, we can assess the feasibility of a larger study based on the proposed method, and get an indication of its generalisability.   
%
We use DBpedia (release 2016-10) in our study as most LOD datasets link to it. We approach this problem as a classification task, using two classification approaches: \textit{alignment-based} (cf. Section~\ref{sec:detourmethods}) and \emph{machine learning-based} (cf. Section~\ref{sec:statisticalmethods}). Since no established procedure exists, we tested different families of methods in an exploratory way. This led us to reuse -- or compare to -- existing work, which provides us with a baseline, which includes Tìpalo \cite{Gangemi2012} as well as other relevant alignments between DBpedia and lexical resources (cf. Section~\ref{sec:detourmethods}).  
%

\subsection{Alignment-based Classification}
\label{sec:detourmethods}
\begin{figure}
	\centering
	\begin{subfigure}[b]{0.47\textwidth}
		\includegraphics[width=\textwidth]{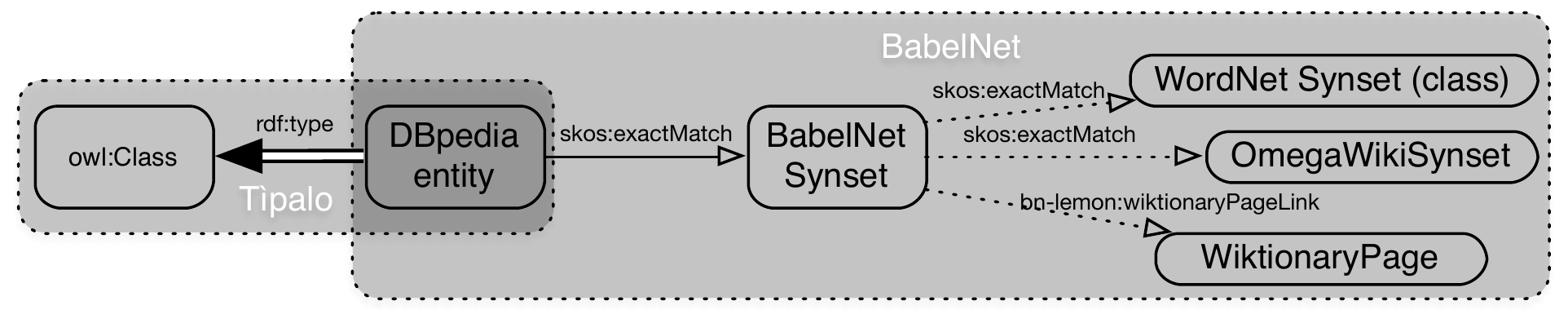}
		\caption{The alignment paths followed by SENECA for selecting candidate classes among DBpedia entities. It identifies as classes all DBpedia entities aligned via BabelNet to a WordNet synset, an OmegaWiki synset or a Wiktionary page, and all DBpedia entities typed as \texttt{owl:Class} in Tìpalo.}
		\label{fig:class-instance}
	\end{subfigure}
	\begin{subfigure}[b]{0.47\textwidth}
		\includegraphics[width=\textwidth]{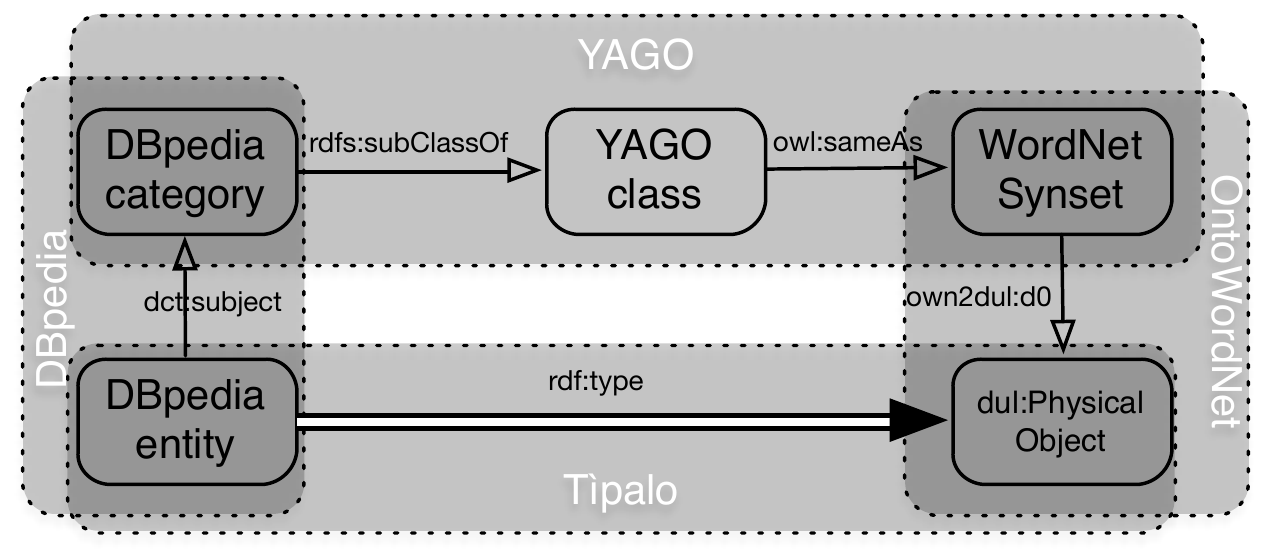}
		\caption{The alignment paths used by SENECA for identifying candidate Physical Objects among DBpedia entities. It navigates the YAGO taxonomy that via OntoWordNet links DBpedia entities to \texttt{dul:PhysicalObject} or Tìpalo that links DBpedia entities to \texttt{dul:PhysicalObject}.}
		\label{fig:physical}
	\end{subfigure}
	\caption{SENECA approach for assessing whether a DBpedia entity is a class or an instance (Figure \ref{fig:class-instance}) and whether it is a physical object or not (Figure \ref{fig:physical}). }
	\label{fig:seneca}
\end{figure}

Alignment-based methods exploit the linking structure of LOD, in particular the alignments between DBpedia, foundational ontologies, and lexical linked data, i.e. LOD datasets that encode lexical/linguistic knowledge. 
%
%
The advantage of these methods is their inherent unsupervised nature.
Their main disadvantages are the need of studying the data models for designing suitable queries, and the potential limited coverage and errors that may accompany the alignments. 
We have developed \textbf{SENECA} (Selecting Entities Exploiting Linguistic Alignments), which relies on existing alignments in LOD, to make an automatic assessment of the foundational distinctions asserted over DBpedia entities. A graphical description of SENECA is depicted in Figure \ref{fig:seneca}.\\
\noindent
\textbf{Class vs. Instance.} As far as this distinction is concerned, SENECA works based on the hypothesis that common nouns are mainly classes and they are expected to be found in dictionaries, while it is less the case for proper nouns, that usually denote instances. This hypothesis was suggested by~\cite{Miller2006}, who manually annotated instances in WordNet, information that SENECA reuses when available. A good quality alignment between the main LOD lexical resources and DBpedia is provided by BabelNet~\cite{Navigli2012}\footnote{We use BabelNet 3.6, which is aligned to WordNet 3.1}. SENECA exploits these alignments and selects all the DBpedia entities that are linked to an entity in WordNet\footnote{\url{http://wordnet-rdf.princeton.edu/}, we use WordNet 3.0 and its alignments to WordNet 3.1, to ensure interoperability with the other resources}, Wiktionary\footnote{\url{https://www.wiktionary.org/}} or OmegaWiki\footnote{\url{http://www.omegawiki.org/}}. With this approach,
63,620 candidate classes have been identified, as opposed to WordNet annotations that only provide 38,701 classes. 
In order to further increase the potential coverage, SENECA leverages the typing axioms of Tipalo~\cite{Gangemi2012}, broadening it to 431,254 total candidate classes. All the other DBpedia entities are assumed to be candidate instances. SENECA criteria for selecting candidate classes among DBpedia entities are depicted in Figure \ref{fig:class-instance}. 

\noindent
\textbf{Physical Object.} Almost 600,000 DBpedia entities are only typed as \texttt{owl:Thing} or not typed at all. However, each DBpedia entity belongs to at least one Wikipedia category. Wikipedia categories have been formalised as a taxonomy of classes (i.e. by means of \texttt{rdfs:subClassOf}) and aligned to WordNet synsets in YAGO~\cite{Suchanek2007}\footnote{We use YAGO 3, aligned to WordNet 3.1}. WordNet synsets are in turn formalised as an OWL ontology in OntoWordNet~\cite{Gangemi2003}\footnote{OntoWordNet is aligned to WordNet 3.0}. OntoWordNet is based on DUL, hence it is possible to navigate the taxonomy up to the DUL class for Physical Object. SENECA looks up the Wikipedia category of a DBpedia entity and follows these alignments. Additionally, it uses Tìpalo, which includes type axioms of DBpedia entities based on DUL classes. SENECA uses these paths of alignments and taxonomical relations, as well as the automated inferences that enable to assess whether a DBpedia entity is a Physical Object or not. 
With this approach, graphically summarised in Figure \ref{fig:physical}, 67,005 entities were selected as candidate physical objects. 

\subsection{Machine Learning-based Classification}
\label{sec:statisticalmethods}
Within machine learning, \textit{classification} is the problem of predicting which category an entity belongs to, given a set of examples, i.e. a training set. 
The training set is processed by an algorithm in order to learn a predictive model based on the observation of a number of \emph{features}, which can be categorical, ordinal, integer-valued or real-valued. 
We have designed our target distinctions in the form of two binary classifications.
We have experimented with eight classification algorithms: J48, Random Forest, REPTree, Naive Bayes, Multinomial Naive Bayes, Support Vector Machines, Logistic Regression, and K-nearest neighbours classifier. We have used WEKA\footnote{\url{https://www.cs.waikato.ac.nz/ml/weka/}} for their implementation.\\
\noindent 
\textbf{Features.} 
The classifiers were trained using the following four features. 

\textbf{Abstract.} Considering that DBpedia entities are all associated with an abstract providing a definitional text, our assumption is that these texts encode useful distinctive patterns. Hence, we retrieve DBpedia entity abstracts, and represent them as 0-1 vectors (bags of words). 
We built a dictionary containing the 1000 most frequent tokens found in all the abstracts of the dataset.
The dictionary is case-sensitive, since the tokens are not normalised.
The resulting vector has a value 1 for each token mentioned in the abstract, 0 for the others.
By inspecting a good amount of abstracts, we noticed that very frequent words, such as conjunctions and determiners, are used in a way that can be informative for this type of classifications.
For example, most of class definitions begin with ``A'' (``A knife is a tool...''). For this reason, we did not remove stop-words.
 
%

\textbf{URI.} We notice that the ID part of URIs is often as informative as a label, and often follows conventions that may be discriminating especially for the class vs. instance classification. In DBpedia, the ID of a URI reflects an entity name (it is common practice in order to make the URI more human-readable), and it always starts with an upper case letter. If the entity's name is a compound term and the entity denotes an instance, each of its components starts with a capital letter. We have also noticed that DBpedia entity names are always mentioned at the beginning of their abstract and, for most of the instance entities, they have the same capitalisation pattern as the URI ID. Moreover, instances tend to have more terms in their ID than classes.
These observations were captured by three numerical features: (i) number of terms in the ID starting with a capital letter, (ii) number of terms in the ID that are also found in the abstract, and (iii) number of terms in the ID.

\textbf{Incoming and Outgoing Properties.} As part of our exploratory approach, we want to test the ability of LOD to show relevant patterns leading to foundational distinctions. Given that triples are the core tool of LOD, we model a feature based on ongoing and outgoing properties of a DBpedia entity. An outgoing property of a DBpedia entity is a property of a triple having the entity as subject. On the contrary, an incoming property is a property of a triple having the entity as object. For example, considering the triple \texttt{dbr:Rome :locatedIn dbr:Italy}, the property \texttt{:locatedIn} is an outgoing property for \texttt{dbr:Rome} and an incoming property for \texttt{dbr:Italy}. For each DBpedia entity, we count its incoming and outgoing properties, per type. For example, properties such as \texttt{dbo:birthPlace} or \texttt{dbo:birthDate} are common outgoing properties of an individual person, hence their presence suggests that the entity is an individual. 	

\textbf{Outcome of SENECA.} Following an exploratory approach, we decided to use the output of SENECA as a binomial feature (taking value ``yes'' or ``no'') for the classifiers (excluding Multinomial Naive Bayes classifier). 
	


\section{Reference Datasets for Classification Experiments}
\label{sec:goldstandard}
In order to perform our experiments and evaluate the results of our approach (cf. Section \ref{sec:methods}), we have created two datasets for each of the foundational distinctions under study: one annotated by experts, and another one annotated by the crowd. In this way we can get an indication whether foundational distinctions match common sense (cf. \emph{RQ1}). The resulting datasets include a sample of annotated DBpedia entities and are available online \footnote{\url{https://github.com/fdistinctions/ijcai18}}. 

\noindent
\textbf{Selecting DBpedia Entities.} The first step to build the datasets is to select a sample of DBpedia entities to be submitted for annotations. It is not straightforward to select a balanced number of classes and instances from DBpedia. A random selection would cause a strong unbalance towards instances because DBpedia contains a larger number of named individuals -- e.g. Rome or Barack Obama -- than concepts. A possible solution could be to manually select a sufficient equal number of DBpedia instances and classes, however this may inject a bias in the datasets. We have opted for a compromise solution by following the intuition that if entities are represented as vectors, their neighbour vectors would include classes and instances with a more balanced ratio than the random choice. As for minimising the bias, we only manually select an arbitrarily small (i.e. 20) number of seeds (equally distributed). We leverage NASARI~\cite{CamachoCollados2016}, a resource of vector representations for BabelNet synsets and Wikipedia entities. For each vector corresponding to a seed entity, we retrieve its 100 nearest neighbours\footnote{We observed that by picking 100 nearest neighbour entities the cosine similarity is always above a threshold of 0.6}.
After cleaning off duplicated entities (e.g. Wikipedia redirects), entities without abstracts, disambiguation pages, etc., we still assessed (through expert annotations) an unbalance towards instances. In the light of a broader usage of the same dataset to also annotate the distinction between physical objects and not physical objects, we enriched the sample with class entities representing physical locations (a good source of physical objects). In order to select only physical location classes from DBpedia, we used the corresponding DBpedia category \texttt{dbc:Places} and the SUN database\footnote{\url{https://groups.csail.mit.edu/vision/SUN/}}, a computer vision-oriented dataset containing a collection of annotated images, covering a large variety of environmental scenes, places, and the objects within. We retrieved DBpedia entities whose labels match SUN locations or that belong to the category \texttt{dbc:Places}, and added a number of them to the sample that would suffice to improve the balance. As a result, a total number of 4502 entities were collected in the newly created dataset.

\noindent
\textbf{CI\textsubscript{E} Dataset: Class vs. Instance Annotation Performed by Experts.}
Two authors of the paper have manually and independently annotated all entities by indicating whether they were instances or classes, using the associated DBpedia abstract as reference description. Their judgements showed an agreement of 93,6\%: they only disagreed on 286 entities. From a joint second inspection, they agreed on additional 281 entities that were initially misclassified by one of the two. Examples of misclassified entities are: \texttt{dbr:Select\_Comfort} (a U.S. manufacturer) that was erroneously annotated as class; \texttt{dbr:Catawba\_Valley\_Pottery} (a kind of pottery) annotated as instance instead of class. Among the remaining entities, five are \textit{polysemous} cases, where the entity and its description point to both types of referents, e.g. \texttt{dbr:Slide\_Away\_Bed} is a trademark commonly used also to refer to a type of beds. 
The authors decided to annotate these entities as classes. As a result, the CI\textsubscript{E} annotated dataset contains 1983 classes and 2519 instances, which is reasonable balanced (44\% classes, 56\% instances).

\noindent
\textbf{CI\textsubscript{C} Dataset: Class vs. Instance Annotation Performed by Crowd.}
The same dataset was then crowdsourced: each worker was asked to indicate whether an entity is a class or an instance based on its name, abstract, and its corresponding Wikipedia article. We want to assess the agreement between the experts and the crowd, which indicates whether foundational distinctions match common sense or not (cf. \emph{RQ1}). The task was executed on Figure Eight \footnote{\url{https://www.figure-eight.com/}} by English speakers with high trustworthiness. The quality of the contributors has been assessed with 51 test questions with a tolerance of only 20\% of errors. 
We collected 22,510 judgments from 117 contributors: each entity was annotated by at least 5 different workers.
For each entity $e$, we computed the level of agreement on each class $c$, weighted by the trustworthiness scores
\begin{equation} \label{eq:agreementOnClass}
	agreement(e, c) = \frac{SumOfTrust(e,c)}{SumOfTrustOfWorkers(e)}
\end{equation}
where $SumOfTrust(e,c)$ is the sum of the trustworthiness scores of the workers that annotated entity $e$ with class $c$; and $SumOfTrustOfWorkers(e)$ is the sum of the trustworthiness scores of all the workers that annotated the entity $e$. Table \ref{tab:crowddataset} reports the results of the task indicating the distribution of classes and instances per level of agreement. The average agreement of the crowd is 95.76\% .
\begin{table}[h]

	\begin{center}
			\resizebox{0.31\textwidth}{!}{
			\begin{tabular}{ c  c  c c}
            \hline
				\textbf{Agreement} &  \textbf{\# Class} &  \textbf{\# Instance} & \textbf{Total} \\\hline
				
				$\ge$ 0.5 & 1934 & 2568 & 4502 \\
				$\ge$ 0.6 & 1884 & 2495 & 4379 \\
				$\ge$ 0.8 & 1631 & 2330 & 3961 \\\hline
				
		\end{tabular}}
	\end{center}
	\caption{\textbf{CI\textsubscript{C} dataset crowd-based annotated dataset of classes and instances.} The table provides an insight of the dataset per level of agreement. Agreement values computed according to Formula ~\ref{eq:agreementOnClass}.}
	\label{tab:crowddataset}
\end{table}
We compared crowd's annotations (with agreement greater than 0.5) against experts' ones.
The judgements of the crowd workers diverge from the experts' only on 193 entities, i.e. agreement is 95,7\%, suggesting that \emph{the instance vs. class foundational distinction matches common sense} (cf. RQ1 applied to this distinction).
Some of those entities (61) also caused a disagreement between experts, hence denoting ambiguous cases.
Examples include polysemic entities such as \texttt{dbr:Zeke\_the\_Wonder\_Dog} or music genres (e.g. \texttt{dbr:Ragga}).

\noindent
\textbf{PO\textsubscript{E} Dataset: Physical Object Annotation Performed by Experts.}
Two authors of the paper further annotated (independently) the dataset by indicating for each entity whether it referred to a physical object (PO) or not (NPO), using its DBpedia abstract as reference description.
They only disagreed on 272 entities, showing an agreement of 93,9\%. 
By means of a joint second inspection, they agreed that the disagreement was caused by errors in the classification, some of which were borderline cases e.g.: \textit{communities} (e.g. \texttt{dbr:Desert\_Lake,\_California}), wrongly interpreted as society instead of neighbourhood, \textit{trademarked materials} (e.g. \texttt{dbr:Waxtite}) and entities with complex description (e.g. \texttt{dbr:Caba{\~{n}}a\_pasiega}).
The resulting PO\textsubscript{E} annotated dataset contains 3055 POs and 1447 NPOs.

\noindent
\textbf{PO\textsubscript{C} Dataset: Physical Object Annotation Performed by Crowd.}
We also crowdsourced the annotations of physical objects vs. not a physical object:
the workers were asked to perform this task by using the entity's name, its abstract, and Wikipedia page as reference descriptions.
The quality of the workers has been assessed with 49 test questions, used to exclude contributors that scored an accuracy lower than 80\%
We collected 25,776 judgments from 173 workers. 
Each entity has been annotated by at least 5 different English speakers. 
Table \ref{tab:physicalobject} summarises the level of agreement associated with the distribution of PO vs. NPO annotations.
The average agreement of the crowd's annotations is 85.48\% .
The agreement between the crowd and the experts is 85,69\%, suggesting that \emph{the PO vs NPO foundational distinction also matches common sense} (cf. RQ1 applied to this distinction).
\begin{table}[h]
	\begin{center}
		\resizebox{0.45\textwidth}{!}{
			\begin{tabular}{ c  c  c c }
            \hline
				\textbf{Agreement} & \textbf{\# Physical Object} & \textbf{\# $\neg$ Physical Object} & \textbf{Total}\\\hline
				
				$\ge$ 0.5 & 3601 & 901 & 4502 \\
                $\ge$ 0.6 & 3448 & 641 & 4089 \\
                $\ge$ 0.8 & 2989 & 335 & 3324 \\\hline
				
		\end{tabular}}
	\end{center}
	\caption{\textbf{PO\textsubscript{C} dataset: crowd-based annotated dataset of physical objects}. The table provides an insight of the dataset per level of agreement. Agreement values computed according to Formula ~\ref{eq:agreementOnClass}.}
	\label{tab:physicalobject}
\end{table}


\section{Experiments Results and Discussion}
\label{sec:results}
\label{sec:experiments}
The results of the performed experiments are expressed in terms of precision, recall and F1 measure, computed for each classification and for each target class (class vs. instance and physical object vs. $\neg$ physical object). The average F1 score is also provided. 
We compare the results of the methods, described in Section \ref{sec:methods}, against the reference datasets CI\textsubscript{E}, CI\textsubscript{C}, PO\textsubscript{E} and PO\textsubscript{C}, described in Section \ref{sec:goldstandard}. As for CI\textsubscript{C} and PO\textsubscript{C}, we only include the annotations having agreement $\ge$ 80\%. 

\subsection{Alignment-based Methods: SENECA}

\noindent
\textbf{Class vs. Instance.}
Table~\ref{tab:detourresultsongeneric} shows SENECA's performance on the class vs. instance classification, by comparing its results with CI\textsubscript{E} and CI\textsubscript{C}.
SENECA shows very good performance with best avg F1 $=$ .836, when compared with CI\textsubscript{C}. Considering that SENECA is unsupervised, and is based on existing alignments in LOD, this result suggests that LOD may better reflect common sense than the expert's perspective, an interesting hint for further investigation on this specific matter.    
\begin{table}[ht]
	\begin{center}
		\resizebox{0.48\textwidth}{!}{
			\begin{tabular}{cccccccc}

	\hline
	\textbf{Dataset}  & \textbf{P\textsuperscript{C}}& \textbf{R\textsuperscript{C}} & \textbf{F\textsupersubscript{C}{1}}  & \textbf{P\textsuperscript{I}}& \textbf{R\textsuperscript{I}} & \textbf{F\textsupersubscript{I}{1}} &  \textbf{\textoverline{F\textsubscript{1}}} \\
    \hline
CI\textsubscript{E} &  .919 & .693& .796 & .753 & .939 & .836 & .813 \\
CI\textsubscript{C} & \textbf{ .935} & \textbf{ .731} & \textbf{ .818} & \textbf{ .778} & \textbf{ .945} & \textbf{ .853} & \textbf{ .836}\\
\hline
 \textbf{Dataset}  &\textbf{P\textsuperscript{PO}}& \textbf{R\textsuperscript{PO}} & \textbf{F\textsupersubscript{PO}{1}}  & \textbf{P\textsuperscript{NPO}}& \textbf{R\textsuperscript{NPO}} & \textbf{F\textsupersubscript{NPO}{1}} &  \textbf{\textoverline{F\textsubscript{1}}} \\
\hline
PO\textsubscript{E} & .877 & .247 & .385 & .561 & .965 & .713 & .548\\
PO\textsubscript{C} & \textbf{.954} & \textbf{.247} & \textbf{.393} & \textbf{.567} & \textbf{.988} & \textbf{.721} & \textbf{.557}\\

\hline
		\end{tabular}}
	\end{center}
	\caption{Results of SENECA on the Class vs. Instance and Physical Object classifications compared against the reference datasets described in Section~\ref{sec:goldstandard}. P\textsuperscript{*}, R\textsuperscript{*} and F\textsupersubscript{*}{1} indicate precision, recall and F1 measure on Class (C), Instance (I), Physical Object (PO) and complement of Physical Object (NPO). \textoverline{F\textsubscript{1}} is the average of the F1 measures.}
	\label{tab:detourresultsongeneric}
\end{table}

\noindent
\textbf{Physical Object.}
Table~\ref{tab:detourresultsongeneric} shows the performance of SENECA on the Physical Object classification task computed by comparing its results with PO\textsubscript{E} and PO\textsubscript{C} (cf. Section \ref{sec:goldstandard}). We observe a significant drop in the best average F1 score (.557) as compared to the class vs. instance classification task (.836). On one hand, this may suggest that the task is harder. On the other hand, the alignment paths followed in the two cases are different, since for classifying Physical Objects more alignment steps are required. In the first case (class vs. instance), BabelNet directly provides the final alignment step (cf. Figure \ref{fig:class-instance}), while in the second case (PO vs. NPO), three more alignment steps are required: DBpedia Category $\rightarrow$ YAGO $\rightarrow$ WordNet (cf. Figure \ref{fig:physical}). It is reasonable to think that this implies a higher potential of error propagation along the flow. This hypothesis is partly supported by \cite{Gangemi2012}, who report a similar drop when they add an automatic disambiguation step followed by an alignment step to DUL classes (including Physical Object). Also for this distinction, SENECA better matches the judgements of the crowd than the experts'.

\subsection{Machine Learning Methods}

We performed a set of experiments with eight classifiers: J48, Random Forest, REPTree, Naive Bayes, Multinomial Naive Bayes, Support Vector Machines, Logistic Regression, and K-Nearest Neighbors (cf. Section \ref{sec:statisticalmethods}). We used a 10-fold cross validation strategy using the reference datasets (cf. Section \ref{sec:goldstandard}). Before training the classifiers, the datasets were adjusted in order to balance the samples of the two classes. The CI\textsubscript{E} and PO\textsubscript{E} datasets were balanced by randomly removing a set of annotated entities. 
CI\textsubscript{C} and PO\textsubscript{C} were balanced by removing entities associated with lower agreement (which constitute weak examples for the classifiers). 
Each classifier was trained and tested with all four features, described in Section \ref{sec:statisticalmethods}, both individually and in all possible combinations, with and without performing feature selection. We found that performing feature selection makes the results worse. 
Having two datasets for each classification (i.e. annotated by the experts and by the crowd) enables multiple configurations of the training set. 
When we train the classifiers with samples from CI\textsubscript{E} and PO\textsubscript{E}, they all have the same weight $=$ 1. Differently, when the samples come from CI\textsubscript{C} and PO\textsubscript{C}, they are weighted according to their associated agreement score \emph{agreement(e, c)}, computed with Formula \ref{eq:agreementOnClass} (cf. Section \ref{sec:goldstandard}). As previously studied by~\cite{Aroyo2015}, this diverse weighing allows a classifier to learn richer information, including ambiguity and consequent entities that may belong to an ``unknown'' class, which better represent human cognitive behaviour. Due to space limits, we only report the results of the best performing algorithm\footnote{The results of all experiments are available online at \url{https://github.com/fdistinctions/ijcai18}}, which is Support Vector Machine, without feature selection, trained and tested on samples associated with an agreement score $\ge$ 80\%. We report on all combinations of features, but \textbf{D} alone (i.e. SENECA's output).

\noindent
\textbf{Class vs. Instance.}
Table~\ref{tab:statisticalresultsongeneric} shows the results of Support Vector Machine, trained on and tested against CI\textsubscript{E} and CI\textsubscript{C}. The best average performance is obtained with CI\textsubscript{C} by combining all features. Combining all features is also the best configuration for each individual classification (i.e. Class (C) and Instance (I)). 
When CI\textsubscript{E} is used there is a slight drop in performance, although the quality of the classification remains high. A possible cause of this result may be the agreement-based weighing provided by CI\textsubscript{C}.

\begin{table*}[ht]
	\begin{center}
		\resizebox{\textwidth}{!}{
			\begin{tabular}{cccc|ccccccc|ccccccc|ccccccc|ccccccc}
            \hline
            \multicolumn{4}{c|}{}  & \multicolumn{7}{c|}{\textit{Results compared against} CI\textsubscript{E}} & \multicolumn{7}{c|}{\textit{Results compared against} CI\textsubscript{C}}  & \multicolumn{7}{c|}{\textit{Results compared against} PO\textsubscript{E}}  & \multicolumn{7}{c}{\textit{Results compared against} PO\textsubscript{C}}\\
            
\textbf{A} & \textbf{U} & \textbf{E} & \textbf{D} & \textbf{P\textsuperscript{C}}& \textbf{R\textsuperscript{C}} & \textbf{F\textsupersubscript{C}{1}}  & \textbf{P\textsuperscript{I}}& \textbf{R\textsuperscript{I}} & \textbf{F\textsupersubscript{I}{1}} &  \textbf{\textoverline{F\textsubscript{1}}} & 
            \textbf{P\textsuperscript{C}}& \textbf{R\textsuperscript{C}} & \textbf{F\textsupersubscript{C}{1}}  & \textbf{P\textsuperscript{I}}& \textbf{R\textsuperscript{I}} & \textbf{F\textsupersubscript{I}{1}} &  \textbf{\textoverline{F\textsubscript{1}}} & 
            \textbf{P\textsuperscript{PO}}& \textbf{R\textsuperscript{PO}} & \textbf{F\textsupersubscript{PO}{1}}  & \textbf{P\textsuperscript{NPO}}& \textbf{R\textsuperscript{NPO}} & \textbf{F
\textsupersubscript{NPO}{1}} &  \textbf{\textoverline{F\textsubscript{1}}} & \textbf{P\textsuperscript{PO}}& \textbf{R\textsuperscript{PO}} & \textbf{F\textsupersubscript{PO}{1}}  & \textbf{P\textsuperscript{NPO}}& \textbf{R\textsuperscript{NPO}} & \textbf{F\textsupersubscript{NPO}{1}} &  \textbf{\textoverline{F\textsubscript{1}}} \\\hline

\checkmark &  &  &  & .927  & .921  & .924  & .921  & .927  & .924  & .924 & .958 & .965 & .961 & .965 & .957 & .961 & .961 & .828 & .814 & .821 & .817 & .832 & .824 & .823 & .879 & .837 & .858 & .844 & .884 & .864 & .861\\

& \checkmark &  &  & .881 & .933  & .906  & .929  & .873  & .909 & .903 & .908 & .970 & .938 & .967 & .902 & .933 & .936 & .615 & .822 & .703 & .732 & .485 & .584 & .644 & .596 & .863 & .705 & .751 & .413 & .533 & .619\\

 &  & \checkmark &  & .854  & .975  & .911  & .971  & .834  & .897  & .904 & .886 & .983 & .932 & .981 & .874 & .924 & .928 & .786 & .865 & .824 & .857 & .764 & .805 & .814 & .782 & .886 & .831 & .868 & .752 & .806 & .818\\

\checkmark & \checkmark &  &  & .928  & .935  & .932  & .935  & .928  & .931  & .932 & .966 & .971 & .968 & .971 & .966 & .968 & .968 & .831 & .829 & .838 & .833 & .832 & .831 & .831& .851 & .811 & .831 & .819 & .857 & .838 & .834\\

\checkmark &  & \checkmark &  & .939  & .943  & .941  & .943  & .939  & .941  & .941 & .971 & .976 & .974 & .976 & .971 & .973 & .974 & .869 & .867 & .868 & .867 & .870 & .868 & .868 & .912 & .853 & .882 & .862 & .918 & .889 & .885\\

\checkmark &  &  & \checkmark & .934 & .927  & .939 & .928  & .934  & .931  & .931 & .966 & .964 & .965 & .964 & .966 & .965 & .965 & .849 & .829 & .835 & .831 & .843 & .837 & .836 & .865 & .834 & .849 & .839 & .869 & .854 & .852\\

  & \checkmark & \checkmark &  & .919  & .968  & .943 & .966  & .914  & .939  & .941 & .961 & .982 & .971 & .981 & .963 & .979 & .971 & .816 & .852 & .833 & .845 & .808 & .826 & .832 & .802 & \textbf{.889} & .842 & .875 & .777 & .823 & .833\\

& \checkmark &  & \checkmark & .881  & .939  & .909  & .935  & .873  & .903  & .906 & .908 & .973 & .939 & .971 & .902 & .935 & .937 & .659 & .761 & .707 & .718 & .607 & .658 & .682 & .951 & .243 & .387 & .565 & \textbf{.987} & .719 & .553\\

  &  & \checkmark & \checkmark & .859  & \textbf{.978}  & .915  & \textbf{.975}  & .846 & .903 & .909 & .889 & \textbf{.987} & .935 & \textbf{.985} & .877 & .928 & .932 & .928 & .735 & .826 & .788 & .943 & .854 & .837 & \textbf{.966} & .762 & .852 & .803 & .973 & .886 & .866\\
      
\checkmark & \checkmark & \checkmark &  & .942  & .946 & .944  & .945  & .942  & .944  & .944 & .973 & .981 & .976 & .980 & .973 & .976 & .976 & .865 & .866 & .866 & .866 & .865 & .865 & .865 & .927 & .847 & .885 & .859 & .933 & .894 & .891\\

\checkmark & \checkmark &  & \checkmark & .939  & .933  & .936  & .934  & .939  & .937  & .936 & .968 & .969 & .968 & .969 & .968 & .968 & .968 & .831 & .824 & .828 & .826 & .833 & .829 & .828 & .878 & .831 & .855 & .838 & .876 & .856 & .853\\

\checkmark &  & \checkmark & \checkmark & .945  & .949 & .943  & .941  & \textbf{.946} & .943  & .943 & .973 & .976 & .975 & .976 & .973 & .975 & .975 & .867 & .862 & .864 & .863 & .868 & .865 & .865 & .922 & .879 & \textbf{.899} & \textbf{.884} & .923 & \textbf{.903} & \textbf{.901}\\

  & \checkmark & \checkmark & \checkmark & .926  & .967  & .946  & .966  & .922  & .944 & .945 & .964 & .981 & .973 & .981 & .964 & .972 & .973 & \textbf{.933} & .736 & .823 & .782 & \textbf{.947} & .857 & .843 & .962 & .759 & .849 & .801 & .975 & .877 & .863\\

\checkmark & \checkmark & \checkmark & \checkmark & \textbf{.946}  & .949 & \textbf{.947}  & .948  & \textbf{.946}  & \textbf{.947}  & \textbf{.947} & \textbf{.981} & .982 & \textbf{.982} & .982 &\textbf{.981} & \textbf{.982} & \textbf{.982}& .871 & \textbf{.877} & \textbf{.871} & \textbf{.879} & .872 & \textbf{.871} & \textbf{.871} & .905 & .866 & .886 & .872 & .909 & .895 & .888\\
\hline

		\end{tabular}}
	\end{center}
	\caption{Results of the Support Vector Machine classifier on Class vs. Instance and Physical Object classification task against the reference datasets described in Section~\ref{sec:goldstandard}. The first four columns indicate the features used by the classifier: \textbf{A} is the abstract, \textbf{U} is the URI, \textbf{E} are incoming and outgoing properties, \textbf{D} are the results of the alignment-based methods. P\textsuperscript{*}, R\textsuperscript{*}, F\textsupersubscript{*}{1} indicate precision, recall and F1 measure on Class (C), Instance (I), Physical Object (PO) and the complement of Physical Object (NPO).  \textoverline{F\textsubscript{1}} is the average of the F1 measures.}
	\label{tab:statisticalresultsongeneric}
\end{table*}

\noindent
\textbf{Physical Object.} 
Table \ref{tab:statisticalresultsongeneric} also shows the results of the Support Vector Machine algorithm trained on and tested against PO\textsubscript{E} and PO\textsubscript{C}. Similarly to the behaviour of SENECA, statistical approaches worsen their overall performance as compared to the case of the class vs. instance classification. We also observe a different behaviour of the individual features. 
The best average performance with PO\textsubscript{E} is achieved by combining all the feature, while the best average performance with PO\textsubscript{C} is achieved by combining the abstract (\textbf{A}), the outgoing/incoming properties (\textbf{E}), and SENECA output (\textbf{D}). In a sense, this confirms that conventions used for creating URI IDs are informative mainly for the class vs. instance distinction.

\subsection{Remarks and Discussion}

We claim that the performed experiments show promising results as far as our research questions are concerned (cf. Section \ref{sec:methods}). Given the diversity and the basic nature of the two distinctions that we have analysed, and the positive results obtained in both cases by applying the same methods with the same configurations, we claim that the proposed methods can be generalised to other foundational distinctions.  \\
\noindent
\emph{RQ1: Do foundational distinctions match common sense?} As anticipated in Section \ref{sec:goldstandard} the high agreement observed among workers that participated in the crowdsourcing tasks, as well as the high agreement between the crowd and the experts, suggest that the foundational distinctions that we have tested do actually match common sense. \\
\noindent
\emph{RQ2: Does the (Semantic) Web provide an empirical basis for supporting foundational distinctions over LOD entities, according to common sense?} We claim that the high average value of F1 measure associated with all experiments indicates that the Web, and in particular LOD, implicitly encodes  foundational distinctions. We also think that, more in general, this is a hint that the Web is a good source for common sense knowledge extraction. We find particularly interesting to observe that the feature \textbf{E} (i.e. ongoing/incoming properties) has always a positive impact, in all features combinations, on the classifier's performance (cf. Table \ref{tab:statisticalresultsongeneric}), for both tasks. This motivates us in conducting further investigations (i) towards identifying and testing additional features based on LOD, e.g. more sophisticated use of assertions and axioms from LOD as well as (ii) to analyse LOD at a much larger scale (e.g. by using LOD Laundromat~\cite{Beek2016}) with an \emph{empirical science} perspective: looking for emerging patterns that may encode relevant pieces of common sense knowledge~\cite{Gangemi2010}. 
Our promising results open a number of possible research directions: besides replicating these experiments at a larger scale, we plan a follow up study concerning the application of the same approach to distinguishing physical objects that can act as locations for other physical objects. This is particularly relevant in order to extract knowledge about where things are usually located in, whether a location is appropriate for an object in terms of its size, etc. Another relevant distinction to be investigated with priority is the one between physical and social objects (e.g. organisations), which is often prone to \emph{systematic polysemy}~\cite{Pustejovsky1998}, i.e. objects that have a same linguistic reference, but different (disjoint) types of referents. For example, the term \emph{National Library} is used to refer both to an organisation (a social object) taking care of the library's collections, and of the related administrative and organisational issues, and to the buildings (physical objects) where the organisational staff works and the collections are located in. Besides covering foundational distinctions, we aim to extend our approach to learn or discover relational knowledge such as the one modelled and encoded in terms of frames~\cite{Fillmore1982,Gangemi2016a}. \\     
\noindent
\emph{RQ3: What ensemble of features, resources, and methods works best to make machines learn foundational distinctions over LOD entities?} 
According to our results, statistical methods perform better than alignment-based methods. We use supervised learning and crowdsourcing to test two very diverse foundational distinctions, both very basic in knowledge representation and foundational ontologies. It emerges that two features show the same ability to positively impact on the methods' performance, for both distinctions: \textbf{A} (a text describing the entity) and \textbf{E} (entity's outgoing and incoming properties). Both features convey the semantic description of an entity: the former by means of natural language, which characterises a huge portion of the web, the latter by means of LOD triples, which characterise the semantic web. Based on these observations, we argue that the method can be generalised, even if each specific distinction may benefit from a specialised extension of the feature set. In our case, the \textbf{U} feature (i.e. URI ID) clearly shows  effectiveness for the class vs. instance rather than for PO vs. NPO. \\
A question is whether DBpedia text is special because of its ``standardised'' style of writing. Our experiments and results do not cover this issue, which needs to be assessed in order to provide a stronger support to our claim of generalisability. A similar doubt can be raised as far as outgoing and incoming links are concerned. DBpedia properties mainly come from infoboxes, which also follow, and are influenced by the standardised way of writing Wikipedia pages. Nevertheless, for this feature we argue that the doubt does not apply, since incoming and outgoing properties include links to and from LOD datasets that are outside DBpedia, hence independent from the standardised content of Wikipedia.

\section{Conclusion}
\label{sec:conclusion}
This study reports a set of experiments for assessing whether the Web, and in particular Linked Open Data, provides an empirical basis to extract foundational distinctions, and if they match common sense. For testing the former, we adopt and compare two approaches, namely alignment-based methods and machine learning methods. For the latter we use crowdsourcing and compare the judgements of the crowd with those of experts'. For both questions we observe promising results and define a method that can be generalised to investigate additional distinctions. We plan experiments on other foundational distinctions (e.g. types of locations, objects that can serve as locations or containers, etc.) and with additional methods.
Our ultimate goal is to advance the state of the art of AI tasks requiring common sense reasoning by designing a methodological framework that enables mass-production of common sense knowledge, and its injection into LOD.

\vspace{0.3cm}
\noindent
\textbf{Acknowledgements.}
This work was partly funded by EU H2020 Project MARIO (grant n. 643808).
The authors thank Aldo Gangemi for his enthusiastic support and precious hints.

\bibliographystyle{named}
\bibliography{stlab2}

\end{document}